\documentclass{article}
\usepackage{ijcai24}
\usepackage{comment}
\usepackage{float}
\usepackage{multirow}
\usepackage{booktabs}
\usepackage{times}
\usepackage{soul}
\usepackage{url}
\usepackage[hidelinks]{hyperref}
\usepackage[utf8]{inputenc}
\usepackage[small]{caption}
\usepackage{graphicx}
\usepackage{amsmath}
\usepackage{amsthm}
\usepackage{amsfonts}
\usepackage{booktabs}
\usepackage{algorithm}
\usepackage{algorithmic}
\usepackage[switch]{lineno}
\usepackage{xcolor}
\usepackage{cleveref}

\title{SOSAE: Self-Organizing Sparse Autoencoders}

\author{
Sarthak Ketanbhai Modi$^1$
\and
Zi Pong Lim$^2$\and
Yushi Cao\footnote{Corresponding Author}$^1$\and
Yupeng Cheng$^1$\and
Yon Shin Teo$^2$ \And
Shang-Wei Lin$^1$\\
\affiliations
$^1$Nanyang Technological University\\
$^2$Continental Automotive Singapore\\
\emails
sarthak002003@gmail.com,
zi.pong.lim@continental-corporation.com,
yushi002@e.ntu.edu.sg,
teoyonshin.continental@gmail.com,
\{yupeng.cheng,shang-wei.lin\}@ntu.edu.sg,
}

\begin{document}

\maketitle
\begin{abstract}
The process of tuning the size of the hidden layers for autoencoders has the benefit of providing optimally compressed representations for the input data. However, such hyper-parameter tuning process would take a lot of computation and time effort with grid search as the default option. In this paper, we introduce the Self-Organization Regularization for Autoencoders that dynamically adapts the dimensionality of the feature space to the optimal size. Inspired by physics concepts, Self-Organizing Sparse AutoEncoder (SOSAE) induces sparsity in feature space in a structured way that permits the truncation of the non-active part of the feature vector without any loss of information. 
This is done by penalizing the autoencoder based on the magnitude and the positional index of the feature vector dimensions, which during training constricts the feature space in both terms. Extensive experiments on various datasets show that our SOSAE can tune the feature space dimensionality up to 130 times lesser Floating-point Operations (FLOPs) than other baselines while maintaining the same quality of tuning and performance.

\end{abstract}

\section{Introduction}
\label{sec:Intro}
\begin{figure}[h]
    \centering
    \includegraphics[width=7.5cm]{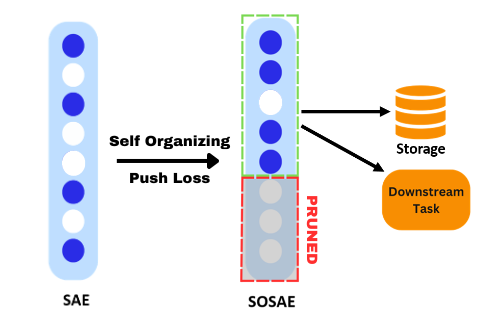}
    \caption{Positional encoding in SOSAE helps to organize the nodes in the feature representations in such a way that the active nodes are at one end of the feature representation hence making it possible to use a shorter length feature representation of the data in the storage or other downstream tasks.}
    \label{fig:SOSAE}
\end{figure}
Autoencoders were first introduced as neural networks that were trained to map their input data to a feature space and reconstruct its input from feature vectors~\cite{doi:10.1126/science.1127647}. Autoencoders have been used for various applications such as data compression~\cite{datacomp,datacomp_2}, data denoising~\cite{denoising,denoising_geophysics}, feature extraction~\cite{supratak2014feature,feature_2} and data generation~\cite{VAE,graphvae,generation_2}, even on large vision datasets with very good results~\cite{pu2016variational}. However, it can become computationally expensive and memory intensive to store the feature representations of all data points in the dataset. 

Introducing sparsity to autoencoders has been used to enhance their representational and generalization capabilities~\cite{SAE_feature,SAE_feature_2} while reducing computation needs~\cite{SAE_feature_3}. Sparsity means to reduce the number of active neurons during training, hence making the representations less densely active. These autoencoders with regularisation that induce sparsity are called Sparse Autoencoders (SAE)~\cite{ng2011sparse}. Various techniques of sparse regularisation such as $L_1$ Loss, $L_2$ Loss~\cite{berthelier2020learning}, and KL-Divergence Loss~\cite{luosparse} were introduced however all of them tend to induce unstructured sparsity which limits the practical savings in terms of computation and space. This is because of irregular memory access which severely limits the exploitation of the sparsity to save computation and memory.

Another way to save computational resources and space whilst achieving a good representation of each data point in the feature space is by tuning the size of the feature representations.
The optimal size of feature representations would depend on the complexity and size of the input data and would be sensitive to the data distribution shift. If the size of the feature representations were to be the same as the input data, the autoencoder would be able to achieve perfect reconstruction and the reconstruction loss would be zero. However, this would be computationally inefficient and lead to suboptimal data compression and feature extraction. However, if the size of the feature representation is too small, the autoencoder would not be able to capture enough features to represent the overall data distribution. 
In addition to input data, the capacity of the encoder and decoder also affects the outcome. Hence these factors make the task of tuning the size of the hidden layer that governs the feature representation, tedious and computationally intensive. 
Some of the existing methods used for the optimization of hidden layer size include Grid Search, Random Search, and Entropy-based algorithms. However, all of these methods~\cite{9287226,KARKKAINEN2023126520} are iterative and hence computationally intensive. They do not attempt to exploit the self-organizing sparsity and excess degree of freedom in the feature space which allows the sparsity to be organized into more desired structures. 

To mitigate the abovementioned issues, we propose a robust and adaptive sparse autoencoder called Self-Organizing Sparse Autoencoder (SOSAE). It simultaneously performs data compression and optimal feature length estimation. 
Self-organizing is a phenomenon observed in  physical~\cite{ebeling2011physics} and biological sciences ~\cite{serugendo2011self} to induce change based on the needs of the environment without any external intervention. This demand-based change without any external force could be beneficial for Neural Networks by helping them adapt their size and number of parameters based on the data. Taking inspiration from this self-organizing, we introduced a "push regularization" function in SOSAE to induce self-organizing behavior into the feature space of the autoencoder, and to drive the nodes with zero activation to higher loss regions, that is towards the end of the feature space. 
The push regularization contains two terms: positional term and magnitude term. The positional term increases exponentially with the position of the dimension in the feature space. Therefore, it encourages the encoder to constrict the feature space in lower dimensions i.e. induce sparsity. Whereas the positional term induces the self organizing behaviour and structures the sparsity.
With the push regularization, non-active nodes are pushed towards the end of the feature space and thus can be pruned automatically to obtain the optimal size of the feature representation, as shown in~\Cref{fig:SOSAE}. 

To demonstrate the effectiveness of our method, we conduct extensive experiments on various popular datasets: MINIST~\cite{deng2012mnist}, CIFAR-10, CIFAR-100~\cite{krizhevsky2009learning}, and Tiny ImageNet~\cite{le2015tiny}.
We evaluate the performances of our approach against various state-of-the-art Autoencoders to see whether SOSAE performs well with the reduced dimensionality. Besides, we also compare SOSAE with other dimensionality search baselines to see if SOSAE is effective in saving computational and memory costs. In summary, our contribution is three-fold:
\begin{itemize}
    \item We introduce a new type of regularization, the push regularization term optimizes the hidden layer dimensionality by modeling the bias-variance tradeoff in terms of compression and feature extraction.
    \item Based on the push regularization term, we introduce the Self-Organizing Sparse Autoencoder (SOSAE), which is capable of optimizing the size of feature representations automatically. This is done simultaneously with the training of the weights of the autoencoder. Hence the length of the feature representation becomes a trainable parameter rather than a fixed one.
    \item Extensive evaluations on various datasets demonstrate that SOSAE is effective in automatically optimizing the dimensionality while maintaining high model performance. More specifically, SOSAE can achieve optimal dimensionality up to 130 times fewer FLOPs than other dimensionality optimization methods.

\end{itemize}

\section{Related Works}



\label{sec:Theory}
\subsection{Autoencoders}

Autoencoders were first introduced in \cite{rumelhart1986parallel} as a neural network that is trained to reconstruct its input, with the main purpose as learning in an unsupervised manner an informative representation of the data. The problem was formally defined in \cite{baldi2012autoencoders}, as
\begin{equation}
\centerline{$ arg min_{G,F} E\left[ L\left( x, G\circ F\left( x \right)\right) \right] $} 
\label{eq:Push Loss}
\end{equation}
to learn the functions $F : \mathbb{R}^{n} \rightarrow \mathbb{R}^{p}$ (encoder) and $G : \mathbb{R}^{p} \rightarrow \mathbb{R}^{n}$ (decoder), where $E$ is the expectation over the distribution of $x$, and $L$ is the reconstruction loss function, which measures the distance between the output of the decoder and the input. 
\subsection{Feature Dimensionality Optimization of Autoencoders}
Undercomplete autoencoders~\cite{baldi2012autoencoders} are the type of autoencoders used for data compression. They have a hidden layer which is of lower size than the input layer. During training, the network will learn to reconstruct the essential features from the original input from the compressed feature representation. The dimensionality of hidden layer is very important in order to obtain optimal feature extraction as well as optimal data compression.
\begin{itemize}
    \item If the selected size of the bottleneck layer is too small, it would lead to high information loss hence leading to poor reconstruction by the decoder.
    \item If the selected size of the bottleneck layer is too large, it would lead to waste of space and computational energy hence leading to suboptimal compression.
\end{itemize}

However, the accuracy exponentially drops at a particular range of compression ratio and before which further decreasing the compression ratio does not further increase the accuracy~\cite{9287226}.
Hence it is essential to choose the correct size within a reasonable allowance that provides us with good compression and high accuracy. In \cite{9287226}, they use a mutual information method to find the hidden layer dimensionality and they validate their result by checking whether the proposed dimension in their model falls in the 'elbow' region of the accuracy vs size curve. Information transfer and data compression are two competing attributes of undercomplete autoencoders
hence finding the right balance between them is necessary to choose the dimensionality of the hidden layer and in turn, the size of the feature representations. 
Another work~\cite{ordway2018autoencoder} utilized random discrete grid search to choose a random set of hyperparameter configurations and builds models in a sequential manner. The hyperparameter dimensions explored included hidden layer configurations, stopping values, input dropout ratios and learning rates. One particular work~\cite{ahmed2022hyperparameter} deployed the ISA technique, which is a gradient-free, and population-based search technique, as a meta-heuristic hyperparameter optimizer to help improve the classification results for road classification models in Intelligent Transportation Systems.
However these methods are iterative in nature and require multiple runs in order to find the optimal size of the feature representation. This constitutes a time-intensive and computationally expensive process. 
To the best of authors knowledge, there has been no work done in a combined methods of solving this optimization problem. What we lack is a method that does not require additional steps and tunes the size parameter whilst training depending on the input data. This would eliminate the need to train multiple models with different parameters. 

\subsection{Sparse Autoencoders (SAE)}
Sparse Autoencoders~\cite{ng2011sparse} have an additional regularisation term in their loss function that deactivates a significant portion of their neurons to facilitate more efficient utilization of memory by limiting the available resources at the network's disposal. It learns novel and interesting features by inducing sparsity (i.e., reducing the number of active neurons) in its feature representation. This limits the network's ability to memorize the data hence forcing the network to learn critical features from the data instead. 

In this work we shall use some of the popular Sparse Autoencoders such as $L_1~Regularization~Loss$, $L_2~Regularization~Loss$ and K-Sparse Autoencoders for benchmarking purposes.
With $L_1~Regularization~Loss$, the autoencoder optimization objective becomes:
\begin{equation}
\centerline{$ arg min_{G,F} E\left[ L\left( x, G\circ F\left( x \right)\right) + \lambda\sum_{i}\left| a_i \right|\right] $} 
\label{eq:L1 Sparse}
\end{equation}
Whereas with $L_2~Regularization~Loss$, the autoencoder optimization objective is then:
\begin{equation}
\centerline{$ arg min_{G,F} E\left[ L\left( x, G\circ F\left( x \right)\right) + \lambda\sum_{i} a_i^2\right] $} 
\label{eq:L2 Sparse}
\end{equation}
where $a_i$ is the activation at the $i$th hidden layer and $i$ iterates over all the hidden activations. 
 Another newer variant of Sparse Autocoders known as the $k$-Sparse Autoencoders~\cite{makhzani2013k}, achieves exact sparsity in the hidden representation by only keeping the $k$ highest activations in the hidden layers.
\section{Methodology}
\begin{figure*}[th]
    \centering
    \includegraphics[width=0.85\linewidth]{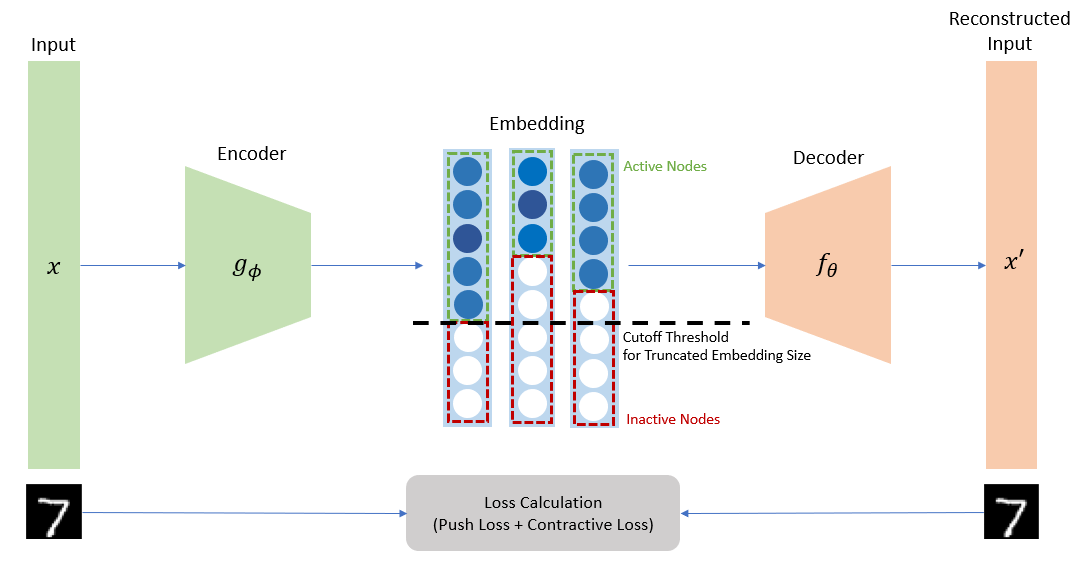}
    \caption{Framework of the SOSAE.}
    \label{fig:framework}
\end{figure*}

As aforementioned, tuning of the optimal size of feature representation can be computationally and time-expensive. Hence, inspired by the self-organization observed in the physical world~\cite{ebeling2011physics}, we propose Self-Organizing Sparse Autoencoder (SOSAE) which induces Self-Organizing property observed in the physical world, into the feature space of autoencoder to provide optimal compression. The Self Organization is induced by the push loss term that organizes the sparsity in the feature representations by "pushing" it to one end hence making it possible to be clipped off. 
%
%
%
\Cref{fig:framework} illustrates the workflow of our SOSAE and shows example feature representations in the hidden layers.
In this framework, the sparsity is organized by the self-organizing induced by the push loss. This structured sparsity is then clipped out without any loss of information and helps us save space and computation in inference. This clipping or truncating is only possible when the sparsity is organized and pushed towards the latter part of the representation or else if the sparsity is unorganized, truncating would lead to information loss. The key difference to note about SOSAE compared to any other method is that it combines the step of hidden layer size selection and data compression together. This save a lot of time. All the other methods of tuning opt for a two step process where step 1 is tuning of the size based on a loss function and iterating through different values until convergence and step 2 is using the model with fixed size of feature representation for data compression. SOSAE combines both these methods by dynamically adjusting the size of the feature representation based on the above mentioned framework.


\subsection{Push Loss}
\noindent

The essence of self organizing in the physical world is the capability to adapt to environmental demands without any external intervention. In the context of neural networks, we seek to induce self-organizing behaviour in the feature space such that the network can adapt to the input data and change the size of its feature representation based on the input data without any external intervention. In this case, external intervention refers to the parameter tuning process. Therefore the network dynamically adapts the size of its feature representation during the training process hence the number of dimensions in the feature space also become a trainable parameter.
To induce the self-organizing behavior in the feature space of neural networks, we propose a push loss term, which contains the positional term and the magnitude term which in this case is the absolute value of the feature representation. This helps to induce the sparsity.
The positional term encodes the positional information of the feature dimensions which helps to push the sparsity. The push loss is defined as:
\begin{align}
    L_N(H) & = \sum_{k=1}^{d_N}(1+\alpha_N)^k \cdot L_{N-1}(H)\\
    L_1(H) &= \sum_{k=1}^{d_1}(1+\alpha_1)^k\cdot |h_{k}|,
    \label{eq:Push Loss}
\end{align}
where $\mathbf{h}$ is the feature representation vector for a particular data point. The $\mathbf{\sum_{k = 1}^{d_1} (1 + \alpha_1)^k \cdot h_i}$ guides the encoder model to induce self-organising behaviour where $(1 + \alpha_1)^k$ holds positional information and $h_i$ hold magnitude information.Here N is the number of dimensions in the feature representation and $d_1, d_2, ..... , d_N$ are the size of the dimensions respectively. $\alpha_1, \alpha_2, ..... , \alpha_N$ follow the condition that:
\[
\alpha_k \propto \frac{\sum_{n = 1}^k d_n}{\sum_{n = 1}^N d_n}
\]
\noindent
In practice, we add an additional regularisation term to the push loss that corresponds to the Frobenius norm of the Jacobian matrix of the encoder activations. This term helps in local space contraction of the feature representation and stabilizes the training while learning more robust features. This additional regularisation term is given by:
\[
\| Jf(x) \|^2 = \sum_{j=1}^{d_h}  \|h_j (1 - h_j) W_{j}\|^2
\]

\subsection{Loss function of SOSAE}
SOSAE is an autoencoder that has self-organizing behaviour in its feature space. The fundamental task of an autoencoder is to reconstruct the input data from samples taken from the feature space. Hence a reconstruction loss is necessary to train the autoencoder.
\begin{equation*}
    \mbox{Reconstruction Loss} = |x - g(f(\hat{x}))|
\end{equation*}
Here $f()$ represents the function modelled by the encoder, $g()$ represents the function modelled by the decoder, $\hat{x}$ is the input data and $x$ is the intended output data.

Apart from this, in order to induce the self-organizing behaviour we add the push loss to this to make the over equation as:
\begin{equation}
  \mathbf{L(x,g(f(\hat{x})))} = \mathbf{|x - g(f(\hat{x}))| +  \sum_{i = 1}^{N} (1 + \alpha)^i \cdot h_i}
\label{eq:SOSAE}
\end{equation}
\subsubsection{Interaction of reconstruction loss and push loss}
The summation of reconstruction loss and push loss helps maintain a balance between the size and information in the feature space. If the push loss term becomes too big, then the neural networks tries to reduce the size of the feature representation in a structured way in order to bring the loss down whereas if the reconstruction loss term becomes too big then the neural network tries to increase the size of the feature representation to ensure adequate information transfer.

\subsubsection{Data Agnostic}
The dynamic size optimization capabilities makes SOSAE data agnostic, since we can use the same architecture for datasets with different levels of complexity or different types of data streams without additional tuning of hidden layer dimensionality. SOSAE can also easily adapt to a dataset with dynamically changing complexity as SOSAE would be able to dynamically change the size of feature representations based on the change in dataset hence eliminating the need for a iterative method to find the hidden layer dimensionality in order to achieve optimal compression without losing information transfer. Due to its combined training process of data compression and size estimation in a single training step, it is able to adapt quickly and efficiently.

\section{Experiments}
\begin{table*}[h]
    \centering
    \small
    \begin{tabular}{@{}p{1.8cm}p{3cm}p{1.5cm}p{1.5cm}p{1.5cm}p{1.5cm}p{1.5cm}p{1.5cm}@{}llll} 
        \toprule
        \textbf{Dataset} & \textbf{Metric} & \textbf{CAE} & \textbf{SOSAE (Ours)}  & \textbf{K-Sparse}& \textbf{Contractive K-sparse}&\textbf{$L_1$ loss }&\textbf{$L_2$ loss}\\
        \midrule
        \multirow{5}{*}{MNIST} & Compressed Length & 400 & \textbf{94}  & 400& 400& 400&400
\\
         & Classification Accuracy & 96.75\% & \textbf{97.58\%}  &\textbf{97.58\%}& 96.26\%& 96.76\%& 96.23\%
\\
        & FLOPs Usage & 100\% & \textbf{23.5\%}  & 25\%& 25\%& 51.7\%&55.6\%
\\
        & Memory Usage & 96 MB & \textbf{22.56 MB}  & 96 MB& 96 MB& 96 MB&96 MB\\
        \midrule
        \multirow{5}{*}{CIFAR-10} & Compressed Length & 800 & \textbf{208}  & 800& 800& 800&800
\\
         & Classification Accuracy & 45.62\% & 46.79\%  & \textbf{46.94 \%}& 45.63\%& 43.06\%&45.24\%
\\
        & FLOPs Usage & 100\% & \textbf{26\%}  & 50 \%& 62.5\%& 71.8\%&77.5\%
\\
        & Memory Usage & 192 MB & \textbf{49.92 MB}  & 192 MB& 192 MB& 192 MB&192 MB\\
         \midrule
        \multirow{3}{*}{CIFAR-100} & Compressed Length & 800 & \textbf{256}  & 800& 800& 800&800\\
         & Classification Accuracy & 18.23\% & \textbf{24.05\%}  & 22.66\%& 19.49\%& 20.33\%&19.35\%\\
        & FLOPs Usage & 100\% & \textbf{32\%}  & 50\%& 62.5\%& 77.5\%&78.8\%\\
        & Memory Usage & 192 MB & \textbf{61.44 MB}  & 192 MB& 192 MB& 192 MB&192 MB\\
        \midrule
        \multirow{3}{*}{Tiny ImageNet} & Compressed Length & 1024 & \textbf{239}  & 1024& 1024& 1024&1024\\
         & Classification Accuracy & 8.26\% & \textbf{8.46\%}& 8.37\%& 7.8\%& 6.88\%&7.35\%\\
        & FLOPs Usage & 100\% & \textbf{23.3\%}  & 39.06\%& 48.8\%& 31.26\%&36.49\%\\
        & Memory Usage & 409.6 MB & \textbf{95.44 MB}  & 409.6 MB& 409.6 MB& 409.6 MB&409.6 MB\\
        \bottomrule
    \end{tabular}
    \caption{Comparisons of SOSAE and various baselines on MNIST, CIFAR-10, CIFAR-100, and Tiny ImageNet datasets.}
    \label{tab:cae_vs_sosae}
\end{table*}
In this section, we empirically show the increase in performance of autoencoders using SOSAE by benchmarking SOSAE against Contractive Autoencoders, Sparse Autoencoders and Denoising Autoencoders for downstream classification task on different datasets. We also analyze the advantage of push loss over non-heuristic and meta-heuristic methods of hidden layer size optimization by analysing the Floating-point Operations (FLOPs) needed for each optimization. Lastly, we compare the denoising capabilities of SOSAE against traditional Denoising Autoencoders. By analyzing these metrics, we demonstrate how SOSAE learns better features, helps in optimal compression even with noisy input data and significantly reduces pretraining time and resources. 
\subsection{Setup}

\subsubsection{Datasets}
We evaluate our methods using four benchmark datasets:
\begin{enumerate}
    \item \textbf{MNIST}~\cite{deng2012mnist}: A dataset of handwritten digits, comprising 70,000  28x28 pixels grayscale images across 10 categories.
    \item \textbf{CIFAR-10} and \textbf{CIFAR-100}~\cite{krizhevsky2009learning}: These datasets contain 60,000 color images each of size 32x32x3 pixels, distributed over 10 and 100 classes respectively, commonly used for image recognition tasks.
    \item \textbf{Tiny ImageNet}~\cite{le2015tiny}: A subset of the ImageNet dataset, Tiny ImageNet includes 100,000 images in 200 classes, scaled down to 64x64 pixels.
\end{enumerate}
These datasets are utilized to assess extract features using the model along with different learning techniques and the extracted feature representations are used to find the accuracy of classification by training a simple single-layer Neural Network.

\subsubsection{Baselines}
Various baselines are used for comparisons, including CAE, SAE, DAE, and embedding size optimization methods. To avoid unfair comparison, following the settings of the CAE paper~\cite{rifai2011contractive}, the size of the hidden layer for the autoencoder is fixed to 400 for MNIST dataset, 800 for CIFAR-10 and CIFAR-100 dataset, and 1024 for Tiny Imagenet. This setting applies to all methods. All baselines are trained with Adam optimizer~\cite{kingma2014adam}.

For all the experiments conducted, the autoencoder model that we use comprises a single fully connected layer as the encoder that takes in data as input and outputs the latent representations, and a decoder which is also a single fully connected layer that takes in the latent representations as input and outputs the reconstructed images. The use of a shallow model is to ensure that the features learned are not entirely due to the capacity of the model but rather due to the loss function and training process used. 





\subsubsection{Metrics}
We adopt the following common metrics for evaluation.
The \textbf{Compressed length} indicates the dimensionality of the feature space, i.e., the size of the hidden layer.\cite{gilbert2024asymmetric} The \textbf{Classification accuracy} indicates the accuracy of the classification task. The \textbf{FLOPs Usage} indicates the percentage of FLOPs used to perform inferences on the feature vectors. This is a key indicator of computational cost.\cite{yu2020auto} The \textbf{Memory Usage} indicates the amount of memory needed to store the feature vectors of the entire dataset.

\subsection{Benchmarking Results}
 
To evaluate the performance of SOSAE with truncated hidden layer size, we perform a comprehensive analysis of SOSAE against CAE, K-sparse, Contractive K-sparse, $L_1$, and $L_2$ sparse autoencoders. The results are shown in~\Cref{tab:cae_vs_sosae},

\paragraph{Compressed length} We observe that SOSAE consistently compressed the input into significantly smaller lengths which are 3.1 to 4.3 times more compression compared to all other variants. This feature directly impacts the memory and computation efficiency during downstream tasks. This significantly more compression suggests that SOSAE is effective in encoding only the most important information. \textbf{The feature lengths for SOSAE (i.e., the size of the hidden layer) are optimized automatically without manual selection.}

\paragraph{Classification Accuracy} SOSAE obtains similar or higher performance against other baseline models in all datasets. More specifically, SOSAE even achieves an up to $5.82\%$ increase in accuracy in MNIST, CIFAR-100, and Tiny ImageNet datasets while maintaining a similar performance in CIFAR-10 dataset.
The results indicate that SOSAE is able to compress the data in the most optimal space without sacrificing learned features or model performance (capacity). We see no drop in performance despite the shorter representation lengths. 


\paragraph{FLOPs Usage} Reduced FLOPs will lead to reduction in time for inference and any other task such as vector search. SOSAE achieves a significant reduction in FLOPs Usage ranging from $23.3\%$ to $32\%$. This is due to the sparsity induced in the representations that save the computation costs. Other methods such as k-sparse, $L_1$, and $L_2$ also provide a reduction in FLOPs Usage however not as much as SOSAE. This is because of the more aggressive sparsity induced in SOSAE due to the exponentially increases loss term with position. 

\paragraph{Memory Usage:} The reduction in memory needed for storage of representations of entire datasets can help in application to various memory-constrained environments. 
SOSAE is the only one that provides these benefits reducing the needed memory by 73.44 MB to 314.16 MB, depending on the dataset. We attribute this to the structured sparsity induced by self-organizing which pushes all the zeroes to one end. These zeroes can be truncated and only the shortened representations are stored. \\

To sum up, SOSAE achieves more efficient compression by optimizing the space and keeping the important features in the representation. More specifically, SOSAE can significantly reduce the computational and memory costs of autoencoders via reducing the sizes of the hidden layers. Such reduction is done via the push loss (without additional tuning) while maintaining high model performance, which highlights the adaptability of the induced self-organizing behavior.
\subsection{Truncation of Length}
\begin{figure}[h]
    \centering
    \includegraphics[width=0.9\linewidth]{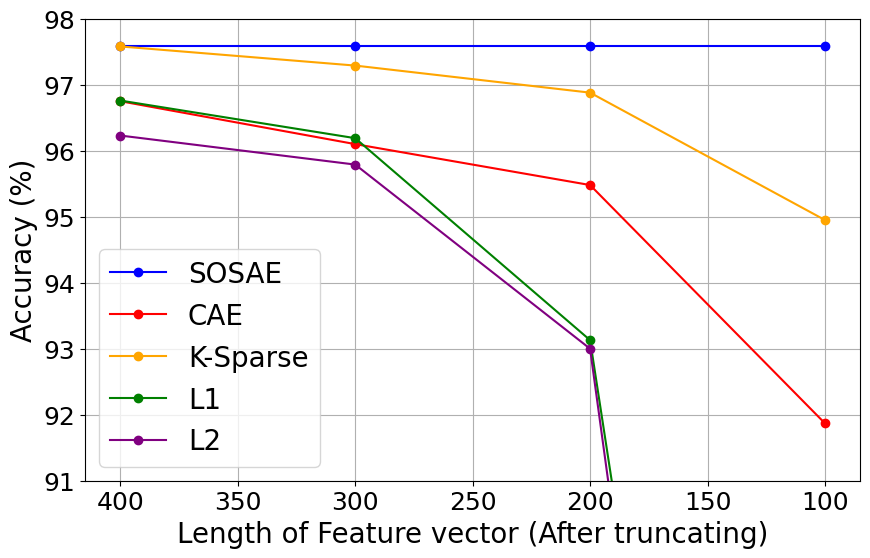}
    \caption{The graph of accuracy plotted against truncated length on MNIST dataset of different variants of autoencoders shows the benefit of self organising the zeroes to the trailing end which allows truncation of feature vector without any loss of information. The k-sparse autoencoder shows some loss of information followed by CAE which shows even more loss of information. $L_1$ and $L_2$ sparsity autoencoders lose most of the information after 200 nodes and the accuracy plummets. }
    \label{fig:truncation}
\end{figure}
\noindent
To validate the robustness of each algorithm against truncation of feature vectors, we gradually remove the tail of the feature vector and compare all methods based on the length of feature vectors as well as 
 the classification accuracy. The result is shown in~\Cref{fig:truncation}. We observe that the feature vectors generated by SOSAE are immune to truncation and the information is not lost even when the last 300 dimensions of the feature vectors are removed. This robustness is not observed in other sparse variants of autoencoders where K-sparse losses some accuracy over truncation, CAE loses even more and $L_1$ and $L_2$ lose almost all information after truncation of 200 dimensions. This shows that even though the other networks have sparse representations however this sparsity cannot be exploited in terms of memory storage. Hence self-organizing of this sparsity is necessary for it to be exploited for memory saving by truncating the zeroes to form a shorter representation.

\subsection{Embedding Size Optimization}
To evaluate the effectiveness of our method, we compare our method against other popular embedding size optimization methods~\cite{9287226,ordway2018autoencoder,ahmed2022hyperparameter}. The results are shown in~\Cref{tab:FLOPS}. 

Results in Table \ref{tab:FLOPS} show that SOSAE uses about 130 times fewer FLOPs for reaching the optimal solution as compared to Grid Search and 34 times fewer FLOPs for reaching the optimal solution as compared to random search. This significantly reduces the time and computation power needed for hyperparameter tuning. Grid search uses a non-heuristic method due to which we need to perform a complete sweep of the search space. Hence we perform 784 interations of training the model and recording the loss in order to evaluate its performance. Finally, we choose the elbow point in the feature representation length vs reconstruction loss which is 88. This ensure the best solution however is very computationally expensive as seen in the table. Random search is meta-heuristic in nature and does not require a complete sweep of the search space. However the quality of the tuning depends on the number of iterations hence in order to get a good quality of tuned parameter, we still require a lot of iterations hence making it cheaper than grid search but still expensive overall. On the other hand, SOSAE dynamically adjusts its length during the training process hence it only requires one-time training. Therefore, the computational costs are reduced significantly. 

\begin{table}[h]
    \centering
    \small
    \renewcommand{\arraystretch}{1.5}
    \begin{tabular}{@{}p{2cm}p{1.75cm}p{0.75cm}p{2cm}@{}} 
        \toprule
        \textbf{Method} & \textbf{Number of Iterations} & \textbf{Tuned Value}&\textbf{Total Number of FLOPs} \\
        \midrule
        Non-Heuristic (Grid Search) & 784 & 88& 8.7 PFLOPs \\
        Meta-Heuristic (Random Search) & 400 &92& 2.27 PFLOPs \\
        SOSAE (Ours) & 1 & 94 & 0.066 PFLOPs \\
        \bottomrule
    \end{tabular}
    \caption{The number of Floating-point operations are calculated for different methods of hyperparameter optimization.}
    \label{tab:FLOPS}
\end{table}

\subsection{Denoising}
\begin{figure*}[h!]
\centering
\label{noises}
\small
\begin{minipage}{6.5cm}
\centering
\includegraphics[width=6cm]{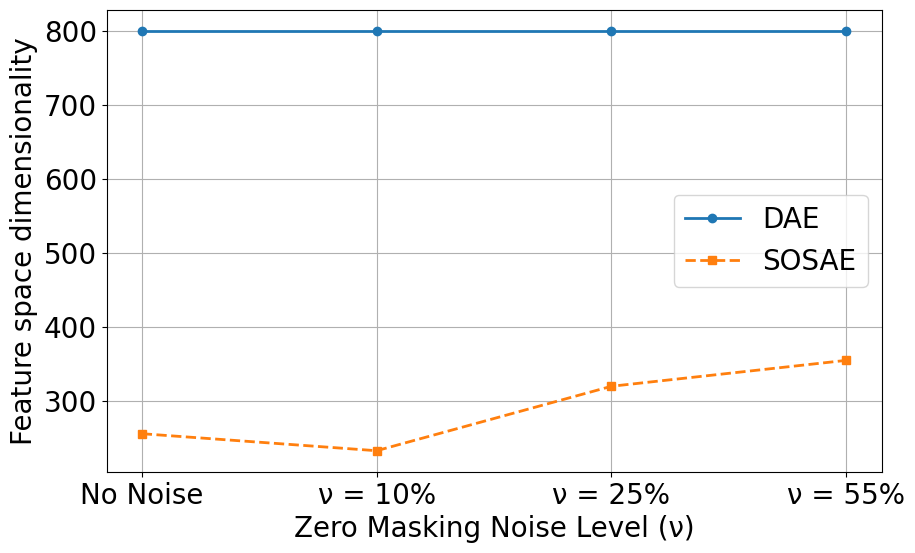}
\centerline{(a) Dimensionality - Zero Masking noise}\medskip
\end{minipage}%
\begin{minipage}{6.5cm}
\centering
\includegraphics[width=6cm]{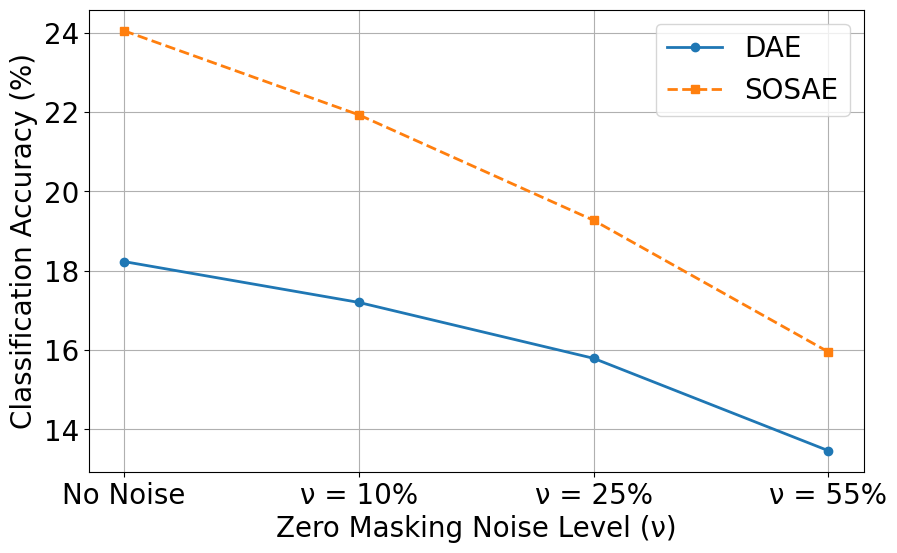}
\centerline{(b) Accuracy - Zero Masking Noise}\medskip
\end{minipage}
\begin{minipage}{6.5cm}
\centering
\includegraphics[width=6cm]{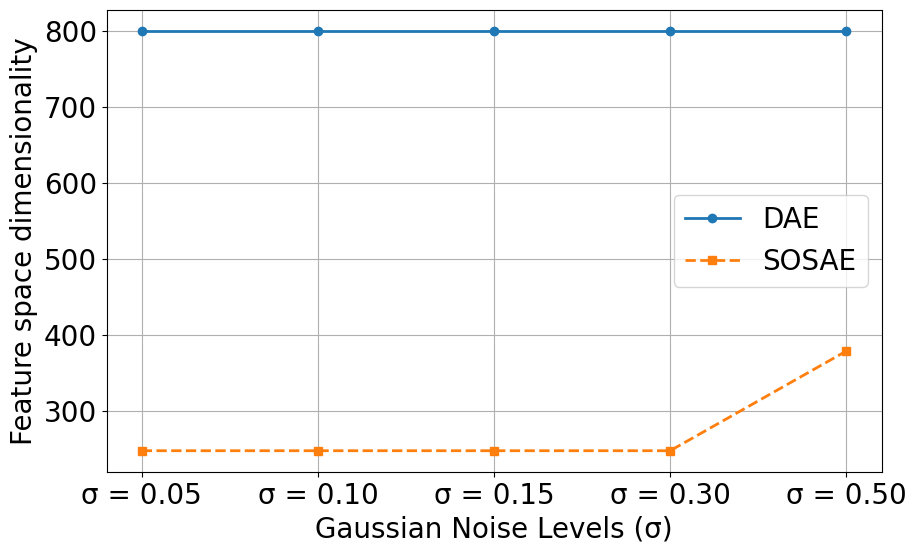}
\centerline{(c) Dimensionality - Gaussian noise}\medskip
\end{minipage}%
\begin{minipage}{6.5cm}
\centering
\includegraphics[width=6cm]{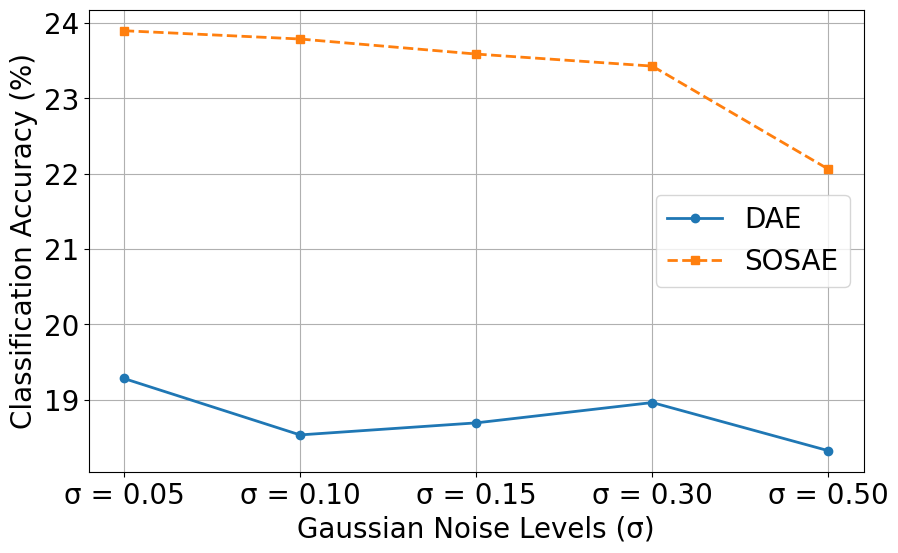}
\centerline{(d) Accuracy - Gaussian Noise}\medskip
\end{minipage}
\caption{Experiments using different levels of Zero Masking and Gaussian noise on SOSAE and DAE}
\label{fig:noise}
\end{figure*}

In this section, we explore the effects of denoising training on SOSAE and benchmark its performance against Denoising Autoencoders. We seek to empirically prove that the dynamic size optimization properties of SOSAE will increase its robustness against noise, hence with noise training it would be able to outperform DAE on downstream classification tasks. We also explore the effect of noise on the compressed length of SOSAE and how it changes with increase in noise levels. 
The considered noises are Zero masking corrupted noise and Gaussian noise~\cite{vincent2010stacked}.

\paragraph{Compressed length:}
Fig \ref{fig:noise} shows the results of SOSAE and DAE on different noise types and levels. For masked noise, we see increasing compressed lengths with an increase in $\nu$ for SOSAE.
This is caused by the loss of information due to a higher masking ratio, which makes feature extraction a more complex task that requires more features to achieve good image reconstruction. However, we still see 2.28 to 2.77 times more compression than traditional Denoising Autoencoders. This increased compression is seen even under Gaussian noise where the length for SOSAE is mostly constant except for very high levels of noise where it increases. SOSAE shows more robustness by retaining the same compressed length from $\sigma$ = 0.05 to $\sigma$ = 0.3. For $\sigma$ = 0.5, we see that it requires a longer representation in order to achieve good reconstruction. Under Gaussian noise, the compression is up to 3.56 times better than DAE. 
\paragraph{Classification Accuracy:}
Figure \ref{fig:noise} shows how robust SOSAE is with noise training compared to DAE. We see that SOSAE significantly outperforms DAE in all different noise types and noise levels.
Moreover, from $\sigma$ = 0.05 to $\sigma$ = 0.30, it is able to achieve almost similar accuracy to that of noise free data, showing that it does not lose performance even with significant noise addition to input data.
Under Zero Masking Noise, it provides 2.5\% - 6.0\% increase in accuracy on downstream classification task which shows that SOSAE is able to learn more robust features than DAE. It shows similar trends under Gaussian noise where it achieves 4.5\% to 5.2\% better accuracy than DAE.

From the experiments shown, we can infer that SOSAE has better robustness against noise, and is able to achieve better results in denoising tasks compared to DAE with much smaller represntations. Hence it not only saves memory and computation in downstream tasks but also learns more information using less space.
\section{Conclusion}
In this work, we propose SOSAE (Self-Organizing Sparse AutoEncoder), a novel autoencoder that achieves optimal feature representation size selection. This is done by introducing a novel push regularization term that induces sparsity and organizes the non-active neurons by "pushing" these non-active neurons to the end of the hidden layer. Such neurons can be pruned and therefore the (near-)optimal hidden layer size can be obtained. Therefore, SOSAE requires much less computing resources. Experimental results on various datasets demonstrate that SOSAE is effective in optimizing the size of feature representation while maintaining the model's performance. It also reduces computational costs and memory usage significantly while maintaining or achieving better performance. 
SOSAE is effective in optimizing the size of hidden layers, demonstrating a notable reduction in computational costs and memory required for storage.
\newpage
\clearpage
\bibliographystyle{named}
\bibliography{IJCAI_DSO/IJCAI_2024/main}
\end{document}